\documentclass[10pt,twocolumn,letterpaper]{article}
\usepackage[pagenumbers]{cvpr} 
\usepackage{graphicx}
\usepackage{amssymb}
\usepackage{booktabs}
\usepackage{algorithm}
\usepackage{amsmath}
\usepackage{algpseudocode}

\usepackage{multirow}
\usepackage[accsupp]{axessibility}
\usepackage{adjustbox}
\usepackage{caption}
\usepackage{makecell}
\usepackage{float}
\usepackage{color}
\usepackage{threeparttable}
\usepackage{lipsum}
\usepackage{url}
\usepackage{enumitem} 

\definecolor{cvprblue}{rgb}{0.21,0.49,0.74}
\usepackage[pagebackref,breaklinks,colorlinks,citecolor=cvprblue]{hyperref}
\def\paperID{***} 
\def\confName{CVPR}
\def\confYear{2026}

\title{Beyond Geometry: Artistic Disparity Synthesis for Immersive 2D-to-3D}

\author{
    Ping Chen$^{1,2}$, Zezhou Chen$^{1,2}$, Xingpeng Zhang$^{5}$, Yanlin Qian$^{3}$, Huan Hu$^{1,2}$, Xiang Liu$^{1,2}$, \\
    Zipeng Wang$^{1,2}$, Xin Wang$^{1,2}$, Zhaoxiang Liu$^{1,2,4^*}$, 
     Kai Wang$^{1,2}$, Shiguo Lian$^{1,2^*}$\\[2mm]
    $^{1}$ Data Science \& Artificial Intelligence Research Institute, China Unicom \\
    $^{2}$ Unicom Data Intelligence, China Unicom,
    $^{3}$ DJI Technology Co.,Ltd. \\
$^{4}$ Beijing Key Laboratory of Science Fiction Audio and Video Intelligent Processing\\
$^{5}$ School of Computer Science and Software Engineering, Southwest Petroleum University\\
    {\tt\small \{chenp181, liuzx178, liansg\}@chinaunicom.cn}
}
\begin{document}
\maketitle
\begin{abstract}
Current 2D-to-3D conversion methods achieve geometric accuracy but are artistically deficient, failing to replicate the immersive and emotionally resonant experience of professional 3D cinema. This is because ``geometric reconstruction" paradigms mistake deliberate artistic intent—such as strategic zero-plane shifts for ``pop-out" effects and local depth sculpting—for data ``noise" or ambiguity. This paper argues for a new paradigm: \textbf{Artistic Disparity Synthesis}, shifting the goal from physically accurate disparity estimation to artistically coherent disparity synthesis. We propose \textbf{Art3D}\footnote{This work was supported by the National Natural Science Foundation of China Enterprise Innovation and Development Joint Fund Project U24B20181}, a preliminary framework exploring this paradigm. Art3D uses a dual-path architecture to decouple global depth parameters (macro-intent) from local artistic effects (visual brushstrokes) and learns from professional 3D film data via indirect supervision. We also introduce a preliminary evaluation method to quantify cinematic alignment. Experiments show our approach demonstrates potential in replicating key local out-of-screen effects and aligning with the global depth styles of cinematic 3D content, laying the groundwork for a new class of artistically-driven conversion tools.
\end{abstract}    
\section{Introduction}
\label{sec:intro}

\begin{figure}[t]
    \centering
    \includegraphics[width=\linewidth]{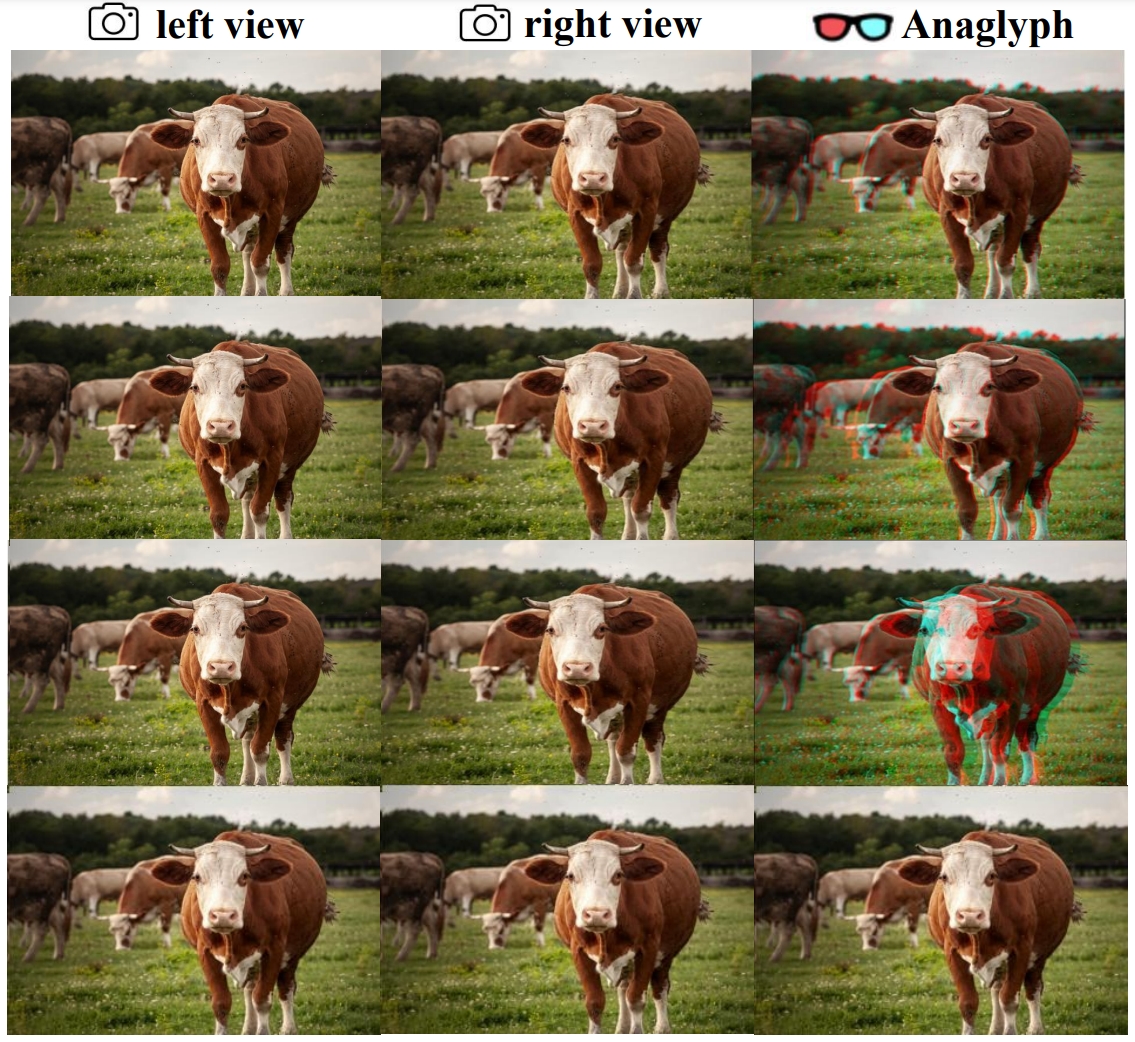}
\caption{
\textbf{Artistic ambiguity in 3D film production.}
The same scene can exhibit distinct 3D perceptions under different artistic directions. 
To illustrate this, we simulate various 3D production choices and visualize their effects. 
From left to right, the three columns show the left view, right view, and the resulting red–cyan anaglyph (best viewed when zoomed in). 
Comparing the first and second rows, variations in the stereo camera’s baseline $b$ and focal length $f$ lead to different disparities, reflecting differences in the \textbf{Mastery of Global Depth}. 
The second and third rows show \textbf{zero-plane ambiguity}, where the perceived depth reference shifts from the foreground cow to the distant forest, often due to creative camera setup or post-production adjustment. 
The fourth row illustrates \textbf{over-simple depth layering}, where disparity remains nearly uniform, likely reflecting cost-efficient rather than artistic intent.
}

    \label{3DA_issues}
\end{figure}

The rapid advancement of virtual reality and immersive media has been driving continuous progress in automated 2D-to-3D conversion technologies \cite{tam20063d, zhang20113d}. State-of-the-art methods, such as diffusion-based approaches have demonstrated remarkable capabilities in generating geometrically coherent and physically plausible stereoscopic effects \cite{zhao2024stereocrafter,geyer2025eye2eye, shvetsova2025m2svid, zhang2024spatialme, diao2024stereo}. From the \textbf{perspective of geometric reconstruction} \cite{garg2016unsupervised, godard2017unsupervised}, they have largely solved the problem of physical correctness. However, when examined through the lens of cinematic art, a critical limitation emerges: these technically proficient conversions still exhibit a significant gap compared to the emotionally resonant, truly immersive experience of professional 3D films. This gap stems not from geometric inaccuracy, but from a systematic absence of artistic intent. In professional 3D film post-production, artistic intent is transmitted through a precise pipeline and ultimately encoded into the disparity map. For example, in \textit{Avatar}'s iconic scene of Jake's first Ikran flight over the Hallelujah Mountains, director James Cameron's artistic goal was to create a ``heart-pounding sensation of soaring flight and grand adventure." To achieve this, the stereoscopic supervisor and stereo artist team executed key operations during post-production: \textbf{(1) Mastery of Global Depth:} They significantly expanded the scene's overall depth budget, creating substantial depth separation between the foreground banshee, the mid-ground floating mountains, and the distant sky background, thereby establishing an epic spatial scale and grandeur.  \textbf{(2) Selection of the Zero-Plane:} They strategically positioned the zero-plane (i.e., the screen plane) within the central region of the mountain range. This choice allowed the banshee and parts of the mountains to appear with negative disparity, seemingly ``bursting out'' of the screen and pulling the audience into the flight path, while the background receded deep into the screen, enhancing the scene's vastness. \textbf{(3)  Sculpting of Local Effects:} For key visual focal points like the tips of the banshee's wings and the character's arms, they performed meticulous local depth enhancement, granting them a more pronounced out-of-screen effect. This acted like the director's ``visual brushstroke,'' precisely guiding audience attention and intensifying the dynamism and immersion of the flight. The ultimate carrier of these artistic decisions is precisely the disparity map \cite{fehn2004depth}. It faithfully encodes all the aforementioned macro parameters and local adjustments, serving as the \textbf{definitive blueprint} for stereoscopic storytelling.

However, the geometric reconstruction paradigm \cite{ideses2007real, po2010automatic, konrad2013learning, xie2016deep3d} dominating automated conversion is fundamentally misaligned with this artistic objective. Methods like MonoDepth \cite{godard2017unsupervised} and MiDaS \cite{ranftl2020towards}, trained to recover physical world accuracy, systematically treat the artistic depth budgets, zero-plane shifts, and local sculpting found in professional 3D as ``noise'' to be suppressed. This leads to a fundamental \textbf{problem of artistic deprivation}: geometrically plausible but narratively impoverished conversions that fail to replicate cinematic-grade immersive experience.

To address this problem of artistic deprivation requires a new research direction beyond pure geometry: Artistic Disparity Synthesis. We posit that the core of next-generation 2D-to-3D conversion should shift from physically accurate disparity estimation'' to\textbf{artistically coherent disparity synthesis},'' thereby establishing a new research paradigm. Our work represents a \textit{preliminary exploration} of this paradigm, aiming to verify the feasibility of generating disparity blueprints that embody both global depth style and local artistic effects. Shifting from physical accuracy to artistic coherence highlights several core challenges that existing technical routes face when confronting artistry: \textbf{(1) Lack of Artistic Supervision:} datasets and loss functions based on physical ground truth cannot provide supervision for learning artistic depth and disparity styles; \textbf{(2) Misinterpretation of Artistic Ambiguity:} disparity variations caused by artistic styles across different films are treated as harmful ambiguity rather than valuable stylistic signals (see Figure \ref{3DA_issues}); \textbf{(3) Systemic Evaluation Mismatch:} existing metrics designed to measure fidelity to physical ground truth systematically penalize the disparity adjustments that constitute artistic expression.

\begin{figure*}[t]
    \centering
    \includegraphics[width=0.9\linewidth]{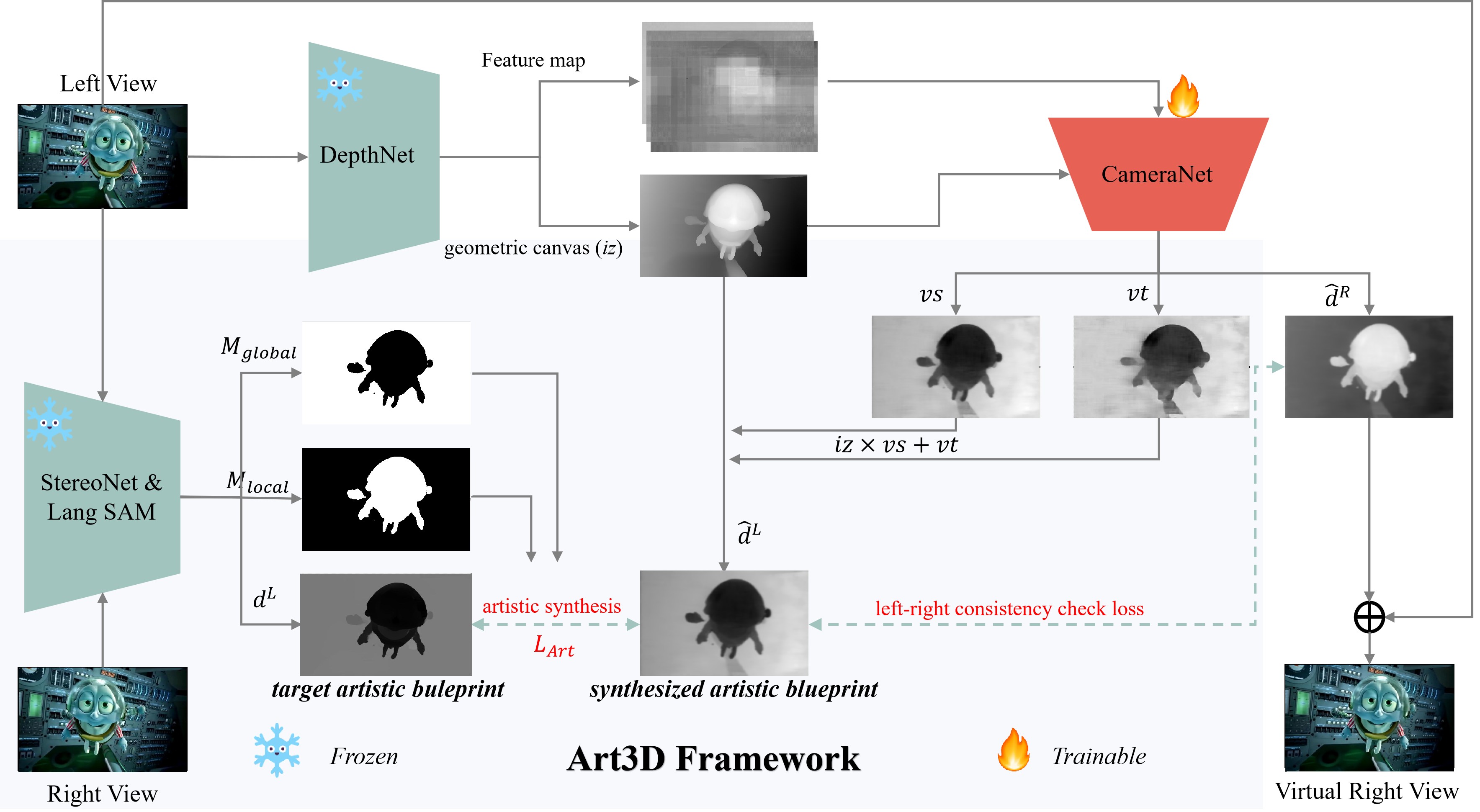}
\caption{
    \textbf{Art3D Framework.} Our pipeline takes a 2D \textit{Left View} as input.
    \textbf{1. Geometric Feature Extraction (Frozen):} `DepthNet' extracts geometric features and the inverse depth map, `geometric canvas ($iz$)'.
    \textbf{2. Artistic Blueprint \& Masks (Frozen, Data Construction Stage):} During the data construction stage (no training or inference required), `StereoNet' estimates the \textit{target artistic blueprint ($d^L$)} and provides a \textit{valid pixel mask ($M_{valid}$)} via its left-right consistency check. Simultaneously, `Lang-SAM' analyzes the \textit{Left View} to generate the \textit{local effects mask} ($M_{local}$). The final \textit{global style mask} ($M_{global}$) is then derived from valid regions that are not part of the local effects (i.e., $M_{valid} \cdot (1-M_{local})$).
    \textbf{3. CameraNet (Trainable):} This network takes features and $iz$ to synthesize the \textit{virtual camera parameters} ($vs, vt$), which are \textbf{pixel-level tensors}, and a preliminary \textit{virtual right disparity map} ($\hat{d}^R$). The virtual left disparity map ($\hat{d}^L$) is then constructed as $iz \times vs + vt$.
    \textbf{4. Dual-Path Supervision:} Our core `$\mathcal{L}_{Art}$' loss guides the learning of $vs, vt$ by comparing $\hat{d}^L$ with $d^L$, weighted by $M_{global}$ (for global style) and $M_{local}$ (for local effects). An additional left-right consistency check loss further refines $\hat{d}^R$.
    \textbf{5. Virtual Right View Synthesis:} The final $\hat{d}^R$, combined with the original \textit{Left View}, can be used to generate the \textit{Virtual Right View} via standard warping techniques.
    For more details on the loss functions, please refer to Sec.\ref{method}.
}
    \label{framework}
\end{figure*}

As a \textit{preliminary exploration} of this new direction, we propose the \textbf{Art3D} framework (see Figure \ref{framework}). Its core attempts to address these challenges through the following approaches:
\begin{itemize}
    \item We design an \textbf{indirect supervision mechanism} that utilizes disparity information extracted from professional 3D films as weak supervisory signals, combined with data filtering strategies, to guide the model in learning global artistic depth styles and local effects.
    \item We introduce a \textbf{dual-path synthesis architecture} that attempts to explicitly separate the learning of stable global parameters representing directorial macro-intent from the learning of local ``artistic brushstrokes" for emphasis.
    \item We construct a \textbf{preliminary evaluation method} that quantifies the performance of generated results against real 3D films in terms of global depth consistency and coverage of local artistic regions, providing a foundation for exploring the quantification of ``artistry."
\end{itemize}
By establishing the disparity map as the carrier of artistic expression and attempting to learn from professional 3D films, the Art3D framework provides a viable approach and foundation for generating more expressive disparity maps. Our experiments indicate that in this preliminary exploration, Art3D demonstrates potential in local out-of-screen effects and shows positive trends in aligning with the \textbf{global depth style} of professional 3D content, laying the initial groundwork for subsequent research in this new paradigm.
\section{Related Work}
\label{sec:related}
We view the evolution of 2D-to-3D conversion and stereoscopic editing as progressing through three main stages: heuristic artistic remapping, learning-based geometric reconstruction, and data-driven artistic style synthesis.

\noindent \textbf{Heuristic Disparity Remapping and Retargeting.}
Early stereoscopic studies explored disparity manipulation beyond strict geometry to achieve artistic and perceptual effects. Representative work introduced non-linear disparity remapping \cite{lang2010nonlinear}, perceptual depth models \cite{didyk2011perceptual}, saliency-aware editing \cite{koppal2010viewer}, and visual comfort optimization \cite{du2013metric, kim2021visual}. Establishing stereoscopic aesthetics via manual disparity editing, these methods require stereo pairs and cannot generalize to monocular inputs.

\noindent \textbf{Geometric Reconstruction Paradigm.}
To handle monocular inputs, most 2D-to-3D methods follow a \textit{geometric reconstruction} paradigm emphasizing physical plausibility. Early work inferred depth from multiple cues to synthesize novel views \cite{zhang20113d, konrad2013learning, mathai2015automatic, calagari2017data}, including motion vectors and block matching \cite{ideses2007real, po2010automatic}, structure-from-motion \cite{liu2015efficient}, and hybrid motion-luminance schemes \cite{zhang2011visual}. Later methods improved robustness via keyframe interpolation \cite{wu2008novel, li2012novel}, skeleton tracking \cite{li2009efficient}, and joint modeling of global gradients and local textures \cite{tsai2011real, yin2015novel}, while learning-based extensions introduced wavelet features, nearest-neighbor regression, and manifold assumptions \cite{ramos2011efficient, konrad2013learning, Herrera2016machine, chowdhary2024efficient}. These developments culminated in early deep models such as Deep3D \cite{xie2016deep3d} and MonDepth \cite{godard2017unsupervised}, leveraging end-to-end training and consistency losses.
Recent work advances this paradigm with diffusion models. The dominant warp-and-inpaint pipeline, used by StereoCrafter \cite{zhao2024stereocrafter}, M2SVid \cite{shvetsova2025m2svid}, and SpatialMe \cite{zhang2024spatialme}, finetunes diffusion models for hole filling after disparity warping. StereoCrafter, trained on 3D movies, normalizes disparity by shifting the zero-plane, discarding the director's original artistic intent. Eye2Eye \cite{geyer2025eye2eye} bypasses warping for direct RGB synthesis and models reflections, but trained on physically accurate VR180 data, its pop-out effect reproduces physical disparity rather than artistic stereoscopic design.
Overall, modern pipelines use foundation models to infer depth or disparity from monocular input and synthesize stereo views via warping and hole filling with perceptual post-processing. Despite industrial practicality, these methods remain geometry-driven and do not model cinematic artistic intent.

\noindent \textbf{Our Paradigm: Data-Driven Artistic Disparity Synthesis.}
While heuristic disparity remapping enables artistic editing for existing stereo pairs and geometric reconstruction provides physically plausible depth from monocular images, our work bridges the gap between them. We argue that strictly physical reconstruction is misaligned with the aesthetic goals of cinematic 3D. Instead, Art3D treats disparity as an artistic carrier and learns cinematic stereoscopic priors from large-scale 3D films, including stereographers’ comfort boundaries accumulated over decades. This enables cross-film 3D style transfer directly from monocular input. By reframing 2D-to-3D from depth estimation to cinematic style modeling, Art3D is, to our knowledge, the first data-driven approach to capture and transfer film-level 3D artistic style, complementing rather than replacing geometric reconstruction.

\section{The Art3D Framework}\label{method}
This section details the \textbf{Art3D} framework (Fig.~\ref{framework}), our proposed solution for \textbf{Artistic Disparity Synthesis}. Our architecture uses a `DepthNet' (for geometry), a `StereoNet' (for the target blueprint), and a core `CameraNet' (for synthesis) to translate the key artistic intents identified in the Sec.\ref{sec:intro}—\textit{Mastery of Global Depth}, \textit{Selection of the Zero-Plane}, and \textit{Sculpting of Local Effects}—into a learnable model. We present our method in three stages: acquiring the geometric input and artistic target, decomposing the artistic signal, and designing the model and loss functions.

\subsection{Geometric Input and Artistic Target}
\label{sec:data}
Our framework learns to map a \textit{Geometric Canvas} (the input) to a \textit{Definitive Blueprint} (the supervision target). These two components are acquired as follows:

\begin{itemize}
    \item \textit{Input Geometric Canvas ($iz$):} We use a robust, pre-trained `DepthNet' (Depth Anything V2  \cite{yang2024depth}) to obtain a stable inverse depth map $iz$ for the left view. This serves as the input geometric foundation upon which our `CameraNet' network will learn to apply artistic edits.
    
    \item \textit{Target Artistic Blueprint ($d^L$):} As stated in Sec.~\ref{sec:intro}, the disparity map serves as the \textit{definitive blueprint} of stereoscopic storytelling. Following recent works~ \cite{geyer2025eye2eye,ranftl2020towards}, we employ an advanced flow-based stereo matcher, `StereoNet' (SEA-RAFT~ \cite{wang2025sea}), to extract both positive and negative disparities $d^L$ from 3D film data, which serves as our target supervision signal.

\end{itemize}

Crucially, this target blueprint $d^L$ is a \textit{mixed signal}: it inherently combines the film's \textit{global depth style} with its \textit{local artistic effects}. To learn from $d^L$ effectively, we must first disentangle these components.

\subsection{Decomposing the Artistic Signal}
\label{sec:decomposing}
We introduce a \textit{dual-path supervision mechanism} to decompose the mixed signal $d^L$. This mechanism is the core of our \textit{dual-path synthesis architecture} and is \textbf{highly robust to errors} from the mask generation tools.

\begin{itemize}
    \item \textit{Local Artistic Mask ($M_{local}$):} To identify regions corresponding to \textit{Sculpting of Local Effects}, we use Lang-SAM  \cite{medeiros2023lsa} with with multiple text prompts (e.g., $``\textit{Non-human out-of-screen object}"$, $``\textit{Foreground character pop-out}"$, etc.). This process does not need to be perfect; it is highly tolerant to \textbf{false negatives}. If a pop-out region is missed by $M_{local}$, it is not detrimental, as it will simply be included in $M_{global}$ and degrade to being supervised by the global style path.

    \item \textit{Global Style Mask ($M_{global}$):} To supervise the global artistic intent (i.e., \textit{Mastery of Global Depth} and \textit{Selection of the Zero-Plane}), we first obtain a set of valid pixels, $M_{valid}$, from `StereoNet`'s left-right consistency matching. The global mask is then defined as the valid pixels excluding the local ones:
        \begin{equation}\label{eq-Gmask}  
        \begin{aligned}  
        M_{global} = M_{valid} \cdot (1-M_{local})
        \end{aligned}  
        \end{equation}
    This process is also highly robust. If `StereoNet' fails to find a match (i.e., $M_{valid}$ has a \textbf{false negative}), our model simply learns the global style from a smaller, sparser subset of pixels for that frame, which effectively acts as a form of data augmentation.
\end{itemize}  

\subsection{Modeling Artistic Synthesis}
\label{sec:modeling}
With our decomposed supervision signals, we now define our dual-path model.

\noindent \textit{Modeling Global Parameters.}
We first model the \textit{global artistic intent} (the signal from $M_{global}$) as a linear transformation of the geometric canvas $iz$, as shown in Eq. \eqref{eq-st}.
\begin{equation}\label{eq-st}  
\begin{aligned}  
d^L = s\cdot iz + t 
\end{aligned}  
\end{equation}
Here, the learnable global parameter $s$ explicitly models the \textit{Mastery of Global Depth} (scaling), and $t$ models the \textit{Selection of the Zero-Plane} (shift).

\noindent \textit{Modeling Local ``Artistic Brushstrokes."}
Eq.~\ref{eq-st} (global-only) cannot capture \textit{Sculpting of Local Effects}; thus, \textbf{Art3D} uses dense per-pixel maps $v_s$ and $v_t$ to synthesize the final artistic blueprint $\hat{d}^L$.
\begin{equation}\label{eq-virtual}  
\begin{aligned}  
\hat{d}^L = vs\cdot iz + vt
\end{aligned}  
\end{equation}
These maps, $vs$ and $vt$, are the learnable parameters that represent the \textit{local ``artistic brushstrokes"} for emphasis, allowing the network to locally refine the disparity on a per-pixel basis.

Our full, unified model (Eq. \ref{eq-v2r}) connects the synthesized blueprint $\hat{d}^L$ to the ground truth blueprint $d^L$ via the residual global parameters:
 \begin{equation}\label{eq-v2r}  
\begin{aligned}  
d^L_k = s \cdot \hat{d}^L_k  + t = s \cdot (vs_k \cdot iz_k + vt_k) + t
\end{aligned}  
\end{equation}
Here, $k$ is the pixel index, and $vs$, $vt$, and $iz$ are tensors of the same dimensions, while $s$ and $t$ are scalars. Since Eq.~\ref{eq-v2r} is an idealization, we estimate $s$ and $t$ in practice by minimizing Eq.~\ref{eq-Art_component}.

\begin{table*}[ht]
\begin{center}
\begin{small}
\caption{Overview of the \textit{CameraNet} Architecture. This is the only component trained in our framework.}
\label{CameraNet}
\begin{tabular}{|c|c|c|}
\hline
\textbf{Operation} & \textbf{Details} & \textbf{Description} \\ \hline
\textbf{Input} & Features and inverse depth (\(iz\)) & Input size = \(512 \times 512\) \\ \hline
\textbf{Downsampling (x3)} & Conv1x1 + ReLU + Conv3x3 + ReLU & Output size = \(64 \times 64\) (encoder features) \\ \hline
\textbf{Upsampling (x3)} & Conv1x1 + ReLU + Deconv3x3 + ReLU & Output size = \(512 \times 512\) (decoder features) \\ \hline
\textbf{Output} & Conv1x1 (out\_channels = 3) & Virtual camera parameters \(vs\) and \(vt\), and right disparity map \(\hat{d}^R\) \\ \hline
\end{tabular}
\end{small}
\end{center}
\end{table*}

\subsection{Loss Function Design}
Our loss function enforces this dual-path learning. It consists of a core synthesis loss to learn the artistic intent and a set of auxiliary losses for geometric consistency.

\noindent \textit{\textbf{Artistic Synthesis Loss ($\mathcal{L}_{Art}$)}}. The core of our framework is the artistic synthesis loss, which is composed of two components, one for each supervision path:
\begin{equation}\label{eq-Art_total}
\begin{aligned}
\mathcal{L}_{Art} = \mathcal{L}_{path}(M_{global}) + \mathcal{L}_{path}(M_{local}) + \mathcal{L}_{st}
\end{aligned}
\end{equation}
Each component loss $\mathcal{L}_{path}(M)$ is defined as the residual of a masked least-squares minimization:
\begin{equation}\label{eq-Art_component}
\begin{aligned}
\mathcal{L}_{path}(M) = \min_{s,t} \sum_{k} M_k \cdot \left\| d^{L}_k - \left( s \cdot \hat{d^{L}_k} + t \right) \right\|^2
\end{aligned}
\end{equation}


Here, $M \in {M_{global}, M_{local}}$, representing the mask used for the corresponding supervision path. This formulation allows the network to learn global parameters (guided by $\mathcal{L}_{path}(M_{global})$) and \textit{local artistic brushstrokes} (guided by $\mathcal{L}_{path}(M_{local})$) simultaneously.

 To ensure that the synthesized disparity $\hat{d}^L$ aligns with the global artistic intent in both overall scale and zero-plane selection, we regularize the global solution ($s, t$) found by the least-squares solver. This encourages $\hat{d}^L$ to directly reflect the global supervision signal, enabling its use during inference without requiring additional $s$ and $t$ estimation.To achieve this property explicitly, we introduce the \textit{Global Style Regularization ($\mathcal{L}_{st}$)}:
\begin{equation}\label{eq-st_loss}  
\begin{aligned}  
\mathcal{L}_{st} = \left\|s-1\right\|^2 + \left\|t\right\|^2 
\end{aligned}  
\end{equation}

\noindent
\textit{\textbf{Auxiliary Loss for Geometric Consistency ($\mathcal{L}_{Aux}$)}}.
To stabilize training and ensure geometric plausibility, we define the auxiliary loss as a weighted sum of smoothness and left-right consistency losses, which are widely used in self-supervised monocular depth estimation  \cite{godard2017unsupervised}:
\begin{equation}\label{eq-aux_loss}
\mathcal{L}_{Aux} = \alpha_1 \mathcal{L}_{grad} + \alpha_2 \mathcal{L}_{lr}
\end{equation}
Here, $\alpha_1$ and $\alpha_2$ are weighting coefficients set according to standard practices in monocular depth estimation \cite{godard2017unsupervised}.
$\mathcal{L}_{grad}$ promotes smoothness in the synthesized disparity $\hat{d}^L$, reducing artifacts, while $\mathcal{L}_{lr}$ enforces consistency between the synthesized left and virtual right disparities, maintaining stereo geometric plausibility.

\noindent \textit{\textbf{Total Loss.}}
The final loss combines the $\mathcal{L}_{Art}$ and $\mathcal{L}_{Aux}$:
\begin{equation}\label{eq-total_loss}  
\mathcal{L} = \mathcal{L}_{Art} + \lambda_1 \mathcal{L}_{Aux}, \quad \lambda_1 = 1 \text{ (default)}.
\end{equation}

\subsection{Evaluation Metrics for Artistic Style}
\label{sec:metrics}
To evaluate our model's ability to synthesize a consistent \textit{global depth style} and \textit{Selection of the Zero-Plane}, standard pixel-wise metrics (e.g., MAE, PSNR) are insufficient, as they measure geometric accuracy, not artistic intent.

We therefore propose an evaluation method to quantitatively assess artistic style. We measure the ``realized" style of any disparity map by fitting it to the geometric canvas $iz$ using the same least-squares minimization defined in our loss function (Eq.~\ref{eq-Art_component}).

This process allows us to compute two distinct style distributions for every image in our test set:
\begin{itemize}
    \item \textit{\textbf{Ground Truth Style $(s_{gt}, t_{gt})$:}} Computed by fitting the true artistic blueprint $d^L$ to $iz$ (using its corresponding validity mask $M_{valid}$).
    
    \item \textit{\textbf{Predicted Style $(s_{pred}, t_{pred})$:}} Computed by fitting the model's synthesized blueprint $\hat{d}^L$ to $iz$ (using $M_{pred}$, a new validity mask generated from the output's own left-right consistency).
\end{itemize}

Our evaluation in Sec.~\ref{sec:exp} then compares the statistical distributions of $(s_{pred}, t_{pred})$ against $(s_{gt}, t_{gt})$ using two statistics:
\begin{itemize}
    \item \textit{ \textbf{Mean ($\mu$):}} Comparing $\mu_{pred}$ to $\mu_{gt}$. This measures the \textit{accuracy} of the learned artistic style. A $\mu_{pred}(s)$ close to $\mu_{gt}(s)$ demonstrates the model has captured the \textit{Mastery of Global Depth} (scaling), while a close $\mu_{pred}(t)$ indicates it has learned the \textit{Selection of the Zero-Plane}.
    
    \item \textit{\textbf{Standard Deviation ($\sigma$):}} Comparing $\sigma_{pred}$ to $\sigma_{gt}$. This is a critical measure of artistic consistency. A high, fluctuating $\sigma$ (unstable $s$ and $t$) creates jarring 3D effects, which can cause viewer discomfort. A low $\sigma_{pred}$ (ideally matching $\sigma_{gt}$) indicates a stable, coherent style crucial for an immersive experience.
\end{itemize}

\section{Experiments}
\label{sec:exp}

\subsection{Experimental Setup}
\noindent \textit{\textbf{Implementation Details.}}
As stated in Sec.~\ref{method}, our 
\textit{DepthNet} and \textit{StereoNet} are frozen and used only for data preprocessing. The only trainable component is \textit{CameraNet} (Fig.~\ref{framework}), a simple U-Net–like model detailed in Table~\ref{CameraNet}. We used the Adam optimizer with a cosine learning rate decay from 0.0005. We trained for 50 epochs on a single A800 80GB GPU with a batch size of 32 and an input size of $512 \times 512$.

\noindent \textit{\textbf{Dataset Construction.}}
As noted in Sec.~\ref{sec:intro} (Fig.~\ref{3DA_issues}, 4th row), we observed that the source quality of public 3D films is highly uneven. In particular, the first sample in the top row of Fig.~\ref{figddciou} shows \textit{over-simple depth layering}, where disparity is nearly uniform. Such frames lack the intended \textit{definitive blueprint} $d^L$, and training on them would mislead the model toward producing flat or low-quality 3D. 

To address this, we propose the Depth-Disparity Consistency IoU (DDC-IoU) to filter out these low-quality frames.
\\
\textbf{\textit{DDC-IoU}}:
We first acquire the geometric canvas $iz$ and artistic blueprint $d^L$ (from Sec.~\ref{sec:data}). We then solve for the global parameters $(s,t)$ that best fit $iz$ to $d^L$ (using the least-squares method from Eq.~\ref{eq-Art_component}) to warp $iz$ into $\hat{d}^{iz}$, a purely geometric disparity map. The DDC-IoU is then defined as:
\begin{equation}\label{eq-ddciou}
\text{DDC-IoU} = \frac{\mathcal{I}(\hat{d}^{iz}) \cap \mathcal{I}(d^L)}{\mathcal{I}(\hat{d}^{iz}) \cup \mathcal{I}(d^L)},
\end{equation}
where $\mathcal{I}(d) = (\nabla_x d) > 0.5$, representing regions of significant disparity change. This metric measures the structural consistency between the geometry ($iz$) and the art ($d^L$).
\\
We applied a $\text{DDC-IoU}\geq0.8$ criterion to select 90,000 high-quality 1080P stereo pairs from 25 well-known 3D movies (e.g., \textit{Hugo}, \textit{The Amazing Spider-Man}, \textit{The Great Gatsby}). Most films are sourced from and follow the dataset protocol described in~ \cite{ranftl2020towards}, with 80,000 pairs used for training and 10,000 for testing.

\textbf{\textit{Local Effects Data}}: Since out-of-screen effects (\textit{Sculpting of Local Effects}) are rare, we manually collected 201 clips from YouTube. After processing, we added ~15,000 frames of out-of-screen data to the training set.

\begin{table}[!]
\caption{Comparison of 2D-to-3D Conversion Paradigms.}
\label{table:paradigm_comparison}
\centering
\begin{adjustbox}{width=0.48\textwidth}
\begin{tabular}{l|c|c}
\hline
\textbf{Method} & \textbf{Global Control (Zero-Plane)} & \textbf{Local Sculpting (Artistic)} \\
\hline
\textit{StereoCrafter} & Manual (Global Shift) & \textbf{No} \\
\textit{Eye2Eye} & Physical (Reproduction) & \textbf{No} \\
\hline
\textbf{Art3D (Ours)} & \textbf{Learned (Global Style)} & \textbf{Yes (Learned)} \\
\hline
\end{tabular}
\end{adjustbox}
\end{table}

\subsection{Ablation Models and SOTA Comparison}
As we are the first to propose this artistic synthesis paradigm, a direct numerical comparison to SOTA methods is not meaningful. Their objectives (geometric/physical fidelity) are fundamentally different from ours (artistic coherence). \textbf{Table \ref{table:paradigm_comparison} provides a high-level conceptual comparison} to clarify these paradigm differences.

\begin{figure*}[t]
    \centering
    \includegraphics[width=\linewidth]{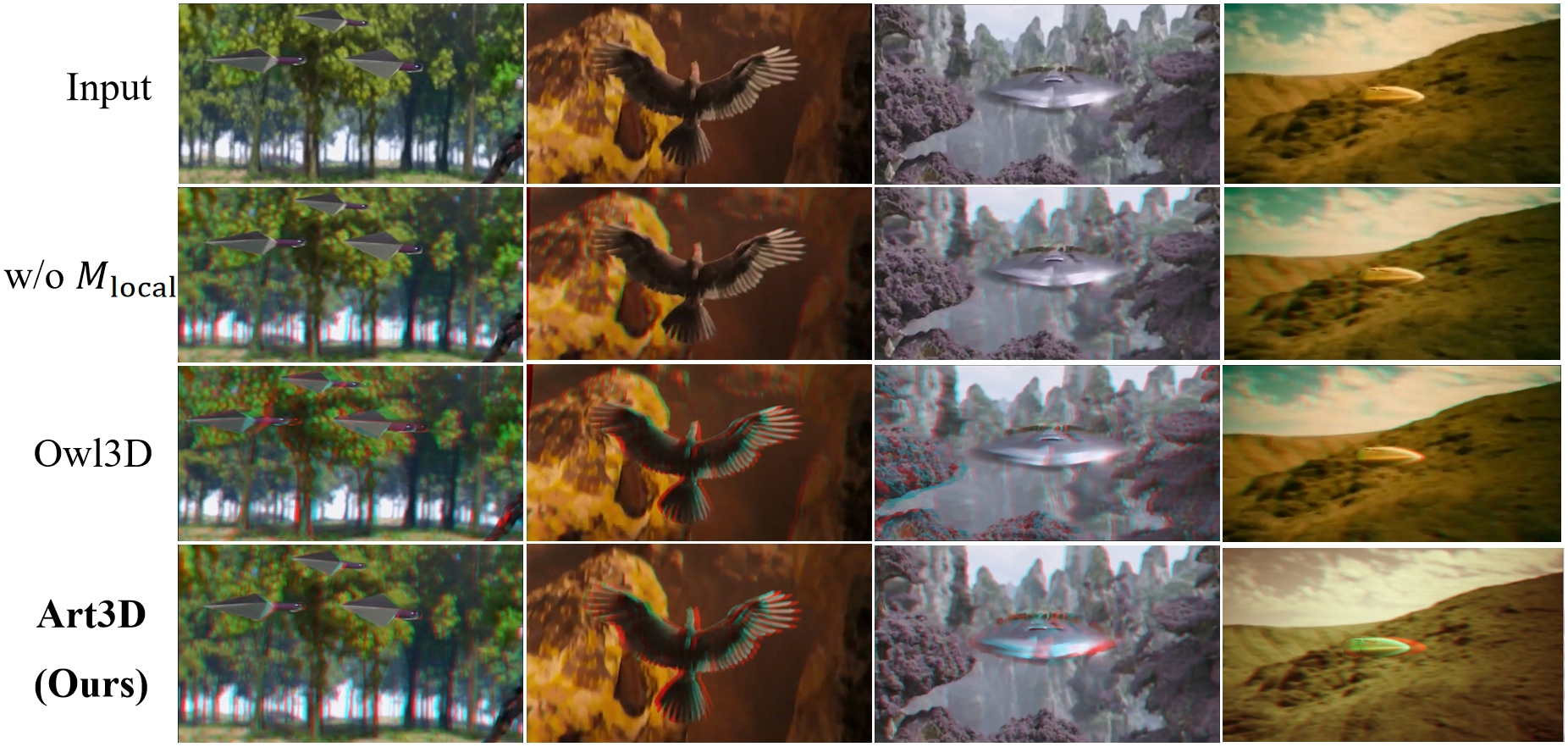}
    \caption{
Qualitative analysis of \textit{Sculpting of Local Effects} on 2D inputs.
\textbf{Row 1:} Input image.
\textbf{Row 2:} Ours (w/o $\mathcal{L}{path}(M_{local})$), which fails to sculpt local pop-out effects.
\textbf{Row 3:} Owl3D, which produces partial effects but lacks artistic consistency.
\textbf{Row 4:} \textbf{Art3D (Full Model)}, which successfully sculpts strong and coherent pop-out effects.
Best viewed zoomed in and with red–cyan anaglyph filters.
}
\label{figpopout}
\end{figure*}

As the table shows, SOTA methods are not designed to learn the \textit{artistic intent} we target. 
As mentioned in Sec.~\ref{sec:related}, \textit{StereoCrafter} \cite{zhao2024stereocrafter} actively \textit{discards} artistic intent by unifying disparity in its data processing. \textit{Eye2Eye} \cite{geyer2025eye2eye}, while able to produce pop-out, learns from \textit{physically-correct} VR180 datasets; its pop-out is a reproduction of \textit{physical} disparity, not the \textit{artistic} (and often non-physical) sculpting our method learns. \textbf{Therefore, our primary evaluation is an ablation study} to validate the effect of our artistic supervision. We compare our full model against a baseline that represents the pure geometric reconstruction paradigm:

\begin{itemize}
    \item \textbf{Baseline (w/o $\mathcal{L}{Art}$)}: Uses only geometric losses ($\mathcal{L}_{Aux}$) with CameraNet + DepthNet, the latter based on the well-established high-precision Depth-Anything-V2, optimized for geometric plausibility but blind to artistic.
   \item \textbf{Art3D (Ours)}: Full model with $\mathcal{L}_{Art}$ + $\mathcal{L}_{Aux}$.
\end{itemize}

\begin{table}[!]
\renewcommand{\arraystretch}{1.0}
\caption{Global Depth Style ($s$) -- Mean (Std) Comparison. A low Std ($\sigma$) indicates high artistic consistency.}
\label{table_s}
\centering
\begin{adjustbox}{width=0.48\textwidth}
\begin{tabular}{l c c c c}
\hline
\textbf{Method} & \textbf{Anime} & \textbf{Sci-Fi}& \textbf{Modern} & \textbf{Mean}\\
\hline
Baseline (w/o $\mathcal{L}_{Art}$) & 0.033(0.022) & 0.030(0.015) & 0.028(0.017) & 0.030(0.018) \\
\textbf{Art3D (Ours)} & \textbf{0.019(0.008)} & \textbf{0.026(0.015)} & \textbf{0.015(0.014)} & \textbf{0.020(0.009)} \\
\hline
\textit{Ground Truth Style} & \textit{0.013(0.010)} & \textit{0.023(0.020)} & \textit{0.012(0.015)} & - \\
\hline
\end{tabular}
\end{adjustbox}
\end{table}

\begin{table}[!]
\renewcommand{\arraystretch}{1.0}
\caption{Zero-Plane Style ($t$) -- Mean (Std) Comparison. A low Std ($\sigma$) indicates high artistic consistency.}
\label{table_t}
\centering
\begin{adjustbox}{width=0.48\textwidth}
\begin{tabular}{l c c c c}
\hline
\textbf{Method} & \textbf{Anime} & \textbf{Sci-Fi}& \textbf{Modern}  & \textbf{Mean}\\
\hline
Baseline (w/o $\mathcal{L}_{Art}$) & 6.96(2.10) & 7.21(2.53) & 6.76(2.43) & 6.98(2.35) \\
\textbf{Art3D (Ours)} & \textbf{6.22(1.05)} & \textbf{6.66(2.01)} & \textbf{5.37(2.93)} & \textbf{6.08(1.80)} \\
\hline
\textit{Ground Truth Style} & \textit{4.35(2.09)} & \textit{5.28(4.68)} & \textit{4.90(3.34)} & - \\
\hline
\end{tabular}
\end{adjustbox}
\end{table}

\subsection{Evaluation of Global Artistic Style}
\label{sec:global_eval}


\noindent \textbf{ \textit{Goal}.} To prove that $\mathcal{L}_{Art}(M_{global})$ successfully teaches the model to learn a \textit{stable and accurate} global artistic style, as defined by the metrics in Sec.~\ref{sec:metrics}.


\noindent \textbf{ \textit{Results}.} Table \ref{table_s} and \ref{table_t} show the results.
The `Baseline (w/o $\mathcal{L}_{Art}$)' model is completely unstable. Its standard deviation ($\sigma$) for both $s$ and $t$ is massive, indicating it is not learning a consistent style, but rather producing random geometric disparities from frame to frame.
In sharp contrast, \textbf{Art3D (Ours)} shows a dramatic reduction in $\sigma$, bringing it much closer to the stable `Ground Truth'. Furthermore, our Mean ($\mu$) values are significantly closer to the `Ground Truth', proving our model has learned the \textit{correct} artistic style (i.e., the correct \textit{Mastery of Global Depth} ($s$) and \textit{Selection of the Zero-Plane} ($t$)).

\begin{figure*}[t]
\centering
\includegraphics[width=0.87\linewidth]{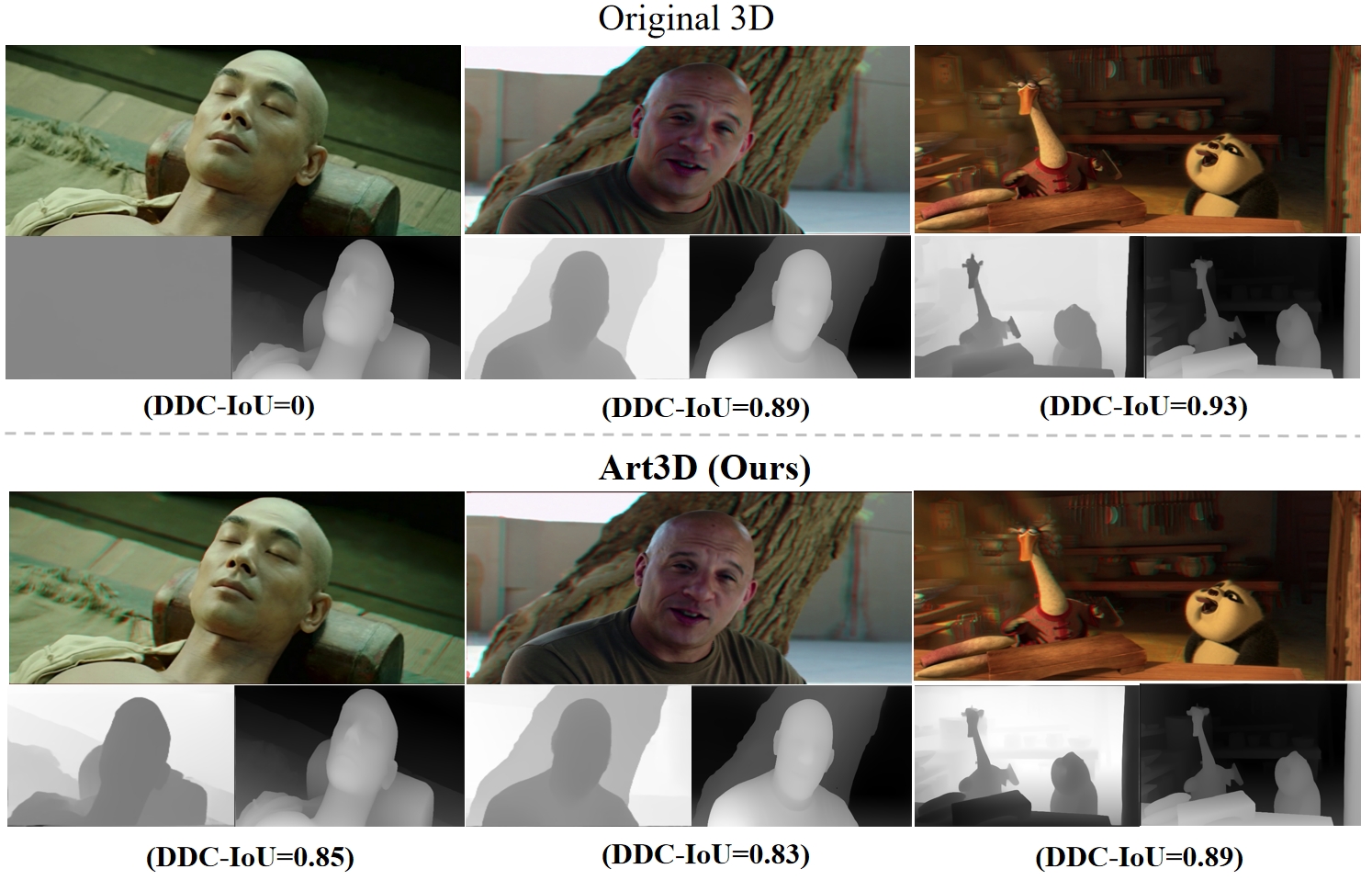}
\caption{\textbf{Analysis of Geometric Consistency via DDC-IoU.} Each row presents three independent samples. For each sample, the Anaglyph 3D view is shown on top. Below it, the corresponding right disparity map ($d^R$ or $\hat{d}^R$) is displayed alongside the right geometric canvas ($iz_R$) derived from that disparity: for ground truth, $iz_R$ is obtained by feeding the true right image into DepthNet; for our estimate, $iz_R$ is computed from the synthesized right view generated by warping the left image using $\hat{d}^R$. \textbf{Top Row (Original 3D Films):} Visualizes the inconsistent quality of source data. The leftmost sample shows poor structural alignment ($\text{DDC-IoU}=0$). The middle and rightmost samples are acceptable ($\text{DDC-IoU} \geq 0.8$). \textbf{Bottom Row (Our Art3D Output):} Shows our model's synthesized blueprints ($\hat{d}^R$) for the \textit{same three scenes} shown above. Our outputs consistently achieve high DDC-IoU scores (e.g., 0.85, 0.83, 0.89), demonstrating that artistic style is learned without corrupting the underlying geometry. Visually, the synthesized disparity maps are coherent and structurally sound.
}
 
\label{figddciou}
\end{figure*}

\subsection{Evaluation of Local Artistic Effects}
\label{sec:local_eval}
\noindent \textbf{\textit{Goal}.} To prove that $\mathcal{L}_{Art}(M_{local})$ effectively teaches the model to perform \textit{Sculpting of Local Effects}.

\noindent  \textbf{\textit{Method}.} We conduct a qualitative comparison on monocular 2D images with strong ``pop-out" potential, none of which have corresponding 3D movie versions, demonstrating our method's generalization to unseen 2D content. We compare three methods: (1) `Ours (w/o $\mathcal{L}_{path}(M_{local})$)', a model trained only on global data; (2) `Owl3D'  \cite{owl3d2024}, a professional 2D-to-3D software; and (3) `Art3D (Full Model)', our complete model.

\noindent \textbf{\textit{Results}.} As shown in Fig.~\ref{figpopout}, the `w/o $\mathcal{L}_{path}(M_{local})$' model (Row 2) produces a flat, geometric-only disparity map. It has learned the global style but \textbf{fails to perform any local sculpting}. The `Owl3D' software (Row 3) can produce some out-of-screen effects (e.g., the dart in the first sample), but its 3D perception is \textbf{inconsistent} across different scenes. In contrast, our \textbf{Art3D} (Row 4) demonstrates both strong and, crucially, \textbf{consistent pop-out effects} across all samples. By viewing the red-cyan details (zoomed in), one can see our model learns the key artistic choice of shifting the zero-plane to the distant background (e.g., mountains, trees) to dramatically enhance the foreground pop-out, proving our dual-path design is essential for learning this \textit{local artistic intent}.

\subsection{Evaluation of Geometric \& Stereo Consistency}

\label{sec:geom_eval}
Having demonstrated that our model learns \textit{artistic style} (Sec.~\ref{sec:global_eval}) and \textit{local effects} (Sec.~\ref{sec:local_eval}), we now verify that our framework also maintains geometric structure and stereo consistency.

\noindent \textit{\textbf{Method.}}
We propose a strong test using our DDC-IoU metric (Eq.~\ref{eq-ddciou}), performed \textit{entirely in the right-view coordinate system}.
First, we generate a Right Geometric Canvas ($iz_R$) by feeding the ground truth right view ($I_R$) into the DepthNet.
Then, we compute the DDC-IoU score between a given right disparity map and this $iz_R$.

This test serves a dual purpose. First, it demonstrates \textit{geometric preservation} by showing the synthesized disparity map is structurally sound and aligned with its target view's geometry. Second, it validates \textit{stereo consistency}: input of our `CameraNet' is the \textit{left} geometric canvas ($iz_L$), but its output is the \textit{right} disparity map ($\hat{d}^R$). By proving that our $\hat{d}^R$ is structurally consistent with the \textit{right} geometric canvas ($iz_R$), we strongly validate that our model has learned a geometrically correct stereo transformation.

\noindent \textit{\textbf{Analysis.}}
Fig.~\ref{figddciou}, which uses the right disparity maps and right geometric canvas described above, provides the visual analysis.
The top row shows original 3D film samples, validating our data filtering strategy. The leftmost sample shows poor structural alignment ($\text{DDC-IoU}=0$), while the middle and rightmost samples are acceptable ($\text{DDC-IoU} \geq 0.8$).
The bottom row shows the corresponding synthesized right disparity maps ($\hat{d}^R$) from our \textbf{Art3D}. Our model's outputs consistently achieve high DDC-IoU scores (e.g., $0.85$, $0.83$, $0.89$), indicating that their structure closely matches the target $iz_R$.
This is a crucial finding. The high DDC-IoU scores validate that our model correctly learns the stereo transformation. This demonstrates that \textbf{in regions without strong artistic intervention} (i.e., non-pop-out areas with small occlusions), our model successfully avoids mode collapse and preserves the underlying geometric structure in the right view, thus producing a coherent stereo blueprint.

\subsection{Artistic Synthesis vs. 2D-to-3D Rendering}
\label{sec:discussion}

This section situates \textbf{Art3D} within the 2D-to-3D pipeline, separating artistic blueprint synthesis from rendering. Our method generates the disparity blueprint $\hat{d}^L$, encoding cinematic decisions while decoupling creative intent from mechanical warping. For end-to-end conversion, this blueprint is used to warp the left view, with occlusions resolved via standard hole-filling \cite{li2025diffueraser,yu2023inpaint} or professional software (e.g., Adobe After Effects). 
\begin{table}[htbp]
\centering
\caption{\small \textbf{User Study}: Art3D vs. Depth-Anything-V2 (Geometric-only, $N=50$ videos, 25 participants)}
\label{tab::user_study}

\resizebox{\linewidth}{!}{
\begin{tabular}{lcccc}
\hline
\textbf{Method} & \textbf{Immersion $\uparrow$} & \textbf{Visual Comfort $\uparrow$} & \textbf{Style Consistency $\uparrow$} & \textbf{Overall Preference $\uparrow$} \\ \hline
Baseline (DA-V2) & 35.2\% & 41.5\% & 22.8\% & 20.0\% \\
\textbf{Art3D (Ours)} & \textbf{64.8\%} & \textbf{58.5\%} & \textbf{77.2\%} & \textbf{80.0\%} \\ \hline
\end{tabular}
}
\end{table}


Therefore, Art3D complements rather than competes with geometric reconstruction. A user study using 50 cinematic clips (avg. 30s) with 25 participants (including professional stereographers and general viewers) comparing Art3D with Depth-Anything-V2 (DA-V2) shows consistent improvements across metrics (Table~\ref{tab::user_study}), with notable gains in Style Consistency (77.2\% vs. 22.8\%) and Overall Preference (80.0\% vs. 20.0\%). These results highlight that geometric estimation alone is insufficient for cinematic immersion, and modeling global and local artistic intent bridges this gap for hybrid pipelines.
\section{Conclusion}
\label{sec:conclusion}
In this paper, we argued that the prevailing \textit{geometric reconstruction} paradigm for 2D-to-3D conversion is misaligned with the goal of cinematic immersion, as it ignores the \textit{artistic intent} (e.g., global depth budget and local sculpting) that defines professional 3D film. We proposed a new paradigm, artistic disparity synthesis, and introduced the \textbf{Art3D} framework to learn this intent. Our core contribution is a \textit{dual-path supervision mechanism}, powered by our $\mathcal{L}_{Art}$ loss, which decomposes the artistic blueprint into \textit{global style} and \textit{local artistic sculpting}, and is highly robust to noisy supervision. Our experiments demonstrated that this approach is highly effective. We proved that our full model learns a stable, coherent global style and successfully synthesizes local "pop-out" effects while maintaining geometric consistency, whereas a baseline without our artistic loss fails. By shifting from physical accuracy to artistic coherence, Art3D offers a first step toward understanding stereoscopic storytelling.
{
     \small
     \bibliographystyle{ieeenat_fullname}
     \bibliography{main}
}


\end{document}